\documentclass[10pt,twocolumn,letterpaper]{article}

\usepackage{wacv}
\usepackage{times}
\usepackage{epsfig}
\usepackage{multirow}
\usepackage{graphicx}
\usepackage{amsmath}
\usepackage{amssymb}

\def\wacvPaperID{0239} 

\wacvfinalcopy 
\ifwacvfinal
\def\assignedStartPage{9876} 
\fi

\ifwacvfinal
\usepackage[breaklinks=true,bookmarks=false]{hyperref}
\else
\usepackage[pagebackref=true,breaklinks=true,colorlinks,bookmarks=false]{hyperref}
\fi

\ifwacvfinal
\else
\fi

\begin{document}

\title{Deep Preset: Blending and Retouching Photos with Color Style Transfer}

\author{Man M. Ho\\
Hosei University\\
Tokyo, Japan\\
{\tt\small man.hominh.6m@stu.hosei.ac.jp}
\and
Jinjia Zhou\\
Hosei University\\
Tokyo, Japan\\
{\tt\small jinjia.zhou.35@hosei.ac.jp}
}

\maketitle

\begin{abstract}
End-users, without knowledge in photography, desire to beautify their photos to have a similar color style as a well-retouched reference. 
However, the definition of style in recent image style transfer works is inappropriate. 
They usually synthesize undesirable results due to transferring exact colors to the wrong destination. It becomes even worse in sensitive cases such as portraits. In this work, we concentrate on learning low-level image transformation, especially color-shifting methods, rather than contextual features matching, then present a novel supervised approach for color style transfer. Furthermore, we propose a color style transfer named Deep Preset designed to 1) generalize the features representing the color transformation from content with natural colors to retouched reference, then blend it into the contextual features of content, 2) predict hyper-parameters (settings or preset) of the applied low-level color transformation methods, 3) stylize content image to have a similar color style as reference. We script Lightroom, a powerful tool in editing photos, to generate 600,000 training samples using 1,200 images from the Flick2K dataset and 500 user-generated presets with 69 settings. Experimental results show that our Deep Preset outperforms the previous works in color style transfer quantitatively and qualitatively. Our work is available at \url{https://minhmanho.github.io/deep_preset/}.
\end{abstract}

\section{Introduction}
Everyone has their life's precious moments captured in photographs. It may tell stories about old memories such as a wedding, a birthday party. Although modern cameras have many techniques to correct the colors and enhance image quality, the natural color style may not express the stories well. Therefore, many powerful tools in editing photos (e.g., Lightroom \cite{lr}) have been released to enrich the preciousness of photographs. However, professional tools require professional skills and knowledge in photography. It causes the end-users difficulty in making their photos prettier, creating an unexpected color style. Motivated by that, many photo applications provide fixed filters to beautify photos conveniently. Unfortunately, the filters are limited and do not meet the user's expectations sometimes. Regularly, experienced users try to mimic a well-retouched photo's color style, giving an overall image of their intended color style. It reveals a correlation between human behavior and the color style transfer task. Inspired by that correlation, we present a supervised approach for color style transfer based on blending and retouching photos. Additionally, we design a specific neural network named Deep Preset to stylize photos and predict the low-level image transformation behind a well-retouched photo.

\begin{figure}[t]
    \centering
    \includegraphics[width=0.9\linewidth]{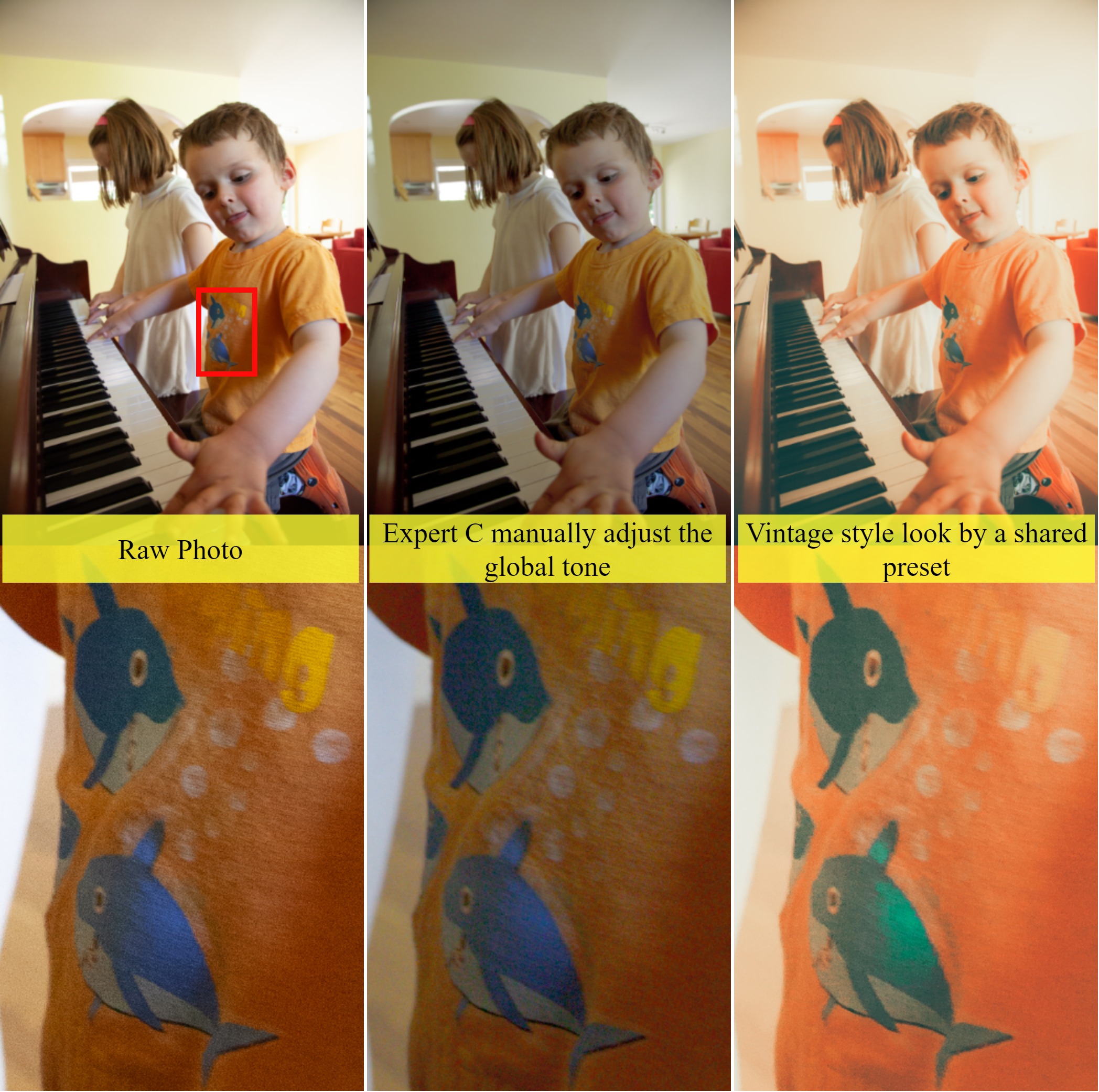}
    \caption{Expert C \cite{bychkovsky2011learning} retouched a raw photo (\textit{left}) to have a better natural-looking (\textit{middle}) using global adjustments. Beyond color correction, the photo can provide better vibes using both global and local adjustments as presets shared over the internet; for example, a vintage style (\textit{right}). It can even change the targeted colors (\textit{the whales}) without distortion creating a novel supervised approach for color style transfer.}
    \label{fig:story-retouch}
\end{figure}

\begin{figure*}[t]
 \centering
 \includegraphics[width=\textwidth]{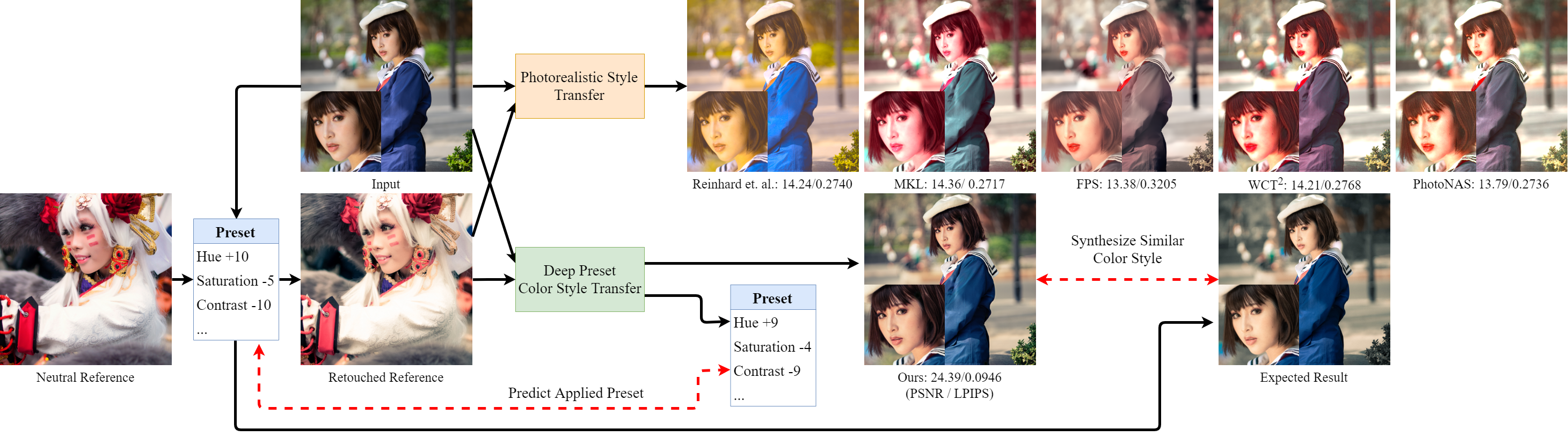}
 \caption{Our overall concept and the problems of previous works Reinhard et. al. \cite{reinhard2001color}, Monge-Kantorovitch Linear (MKL) \cite{pitie2007linear}, Fast Photo Style (FPS) \cite{li2018closed}, WCT$^2$ \cite{yoo2019photorealistic}, PhotoNAS \cite{an2019ultrafast} shown with PSNR$\uparrow$ / LPIPS$\downarrow$ \cite{zhang2018unreasonable}. ($\uparrow$: higher is better, $\downarrow$: lower is better).}
 \label{fig:story}
\end{figure*}

\textbf{Color correctness and blending}. Nowadays, most digital cameras have post-processing techniques to address the shortcomings of the sensor. It provides natural colors for the captured pictures before saving. Unfortunately, the raw photos may not fully show what human eyes actually observe due to a lack of light exposure, low-cost devices, etc. Experienced users then use professional tools to manually correct the global toner (e.g., lightness, white balance, contrast, etc.). Hence, the users' photo adjustments in correcting colors become valuable to the community wishing computers can replace humans to handle it automatically. It motivates Bychkovsky et al. \cite{bychkovsky2011learning} to create a high-quality reference dataset MIT-Adobe FiveK by asking five photography students to retouch raw images. Their work opens a supervised approach for correcting colors and predicting a reasonable photo adjustment. Thanks to the development of deep learning, the problems in automatic color correction are handled surprisingly well. As an example, Afifi et. al. proposed methods to correct exposure \cite{afifi2020learning} and white-balance \cite{afifi2020deepWB}. The Exposure of Hu et al. \cite{hu2018exposure} directly predicts a set of photo adjustment operations retouching big size photos in real-time. Additionally, their work provides users a scheme being able to adjust the photo afterward. Since the mentioned works are color correction for raw images, they only adjust global toner to avoid local changes retaining the original and consistency of taken colors, providing a naturalness. In our work, beyond color correction and global adjustments, we also exploit other blending styles and local adjustments, presenting a novel scheme for color style transfer.
In reality, besides the style of naturalness, the experienced users blend the colors according to their purposes using not only global adjustment (e.g., Brightness, Contrast, Saturation, etc.) but also local adjustments (e.g., Red Hue Shifting, Blue Saturation Adjustment). For example, a memorial photo is retouched for a vintage style telling an old story with a nostalgic vibe; besides, the local adjustment helps to change the blue colors of \textit{the whales} to \textit{green ones}, as shown in Figure \ref{fig:story-retouch}. After done photo adjusting, all adjustments will be stored as settings (preset) representing low-level color transformation methods. Usually, on the internet, a preset is shared with a photo retouched by that preset, helping end-users without photography knowledge to understand the color style before using. It opens a novel scheme in generating photos having the homologous color style for training. The ill-posed problem is how to generalize features representing color transformation, even when the image content is different, and build an efficient color style transfer. In this work, we consider two approaches: 1) Let the proposed Deep Preset predict the applied preset, which is photo adjustments behind the retouched reference. The extracted features from various image contents thus define the transformation. However, predicting an accurate preset is a difficult issue; we thus 2) add a Positive Pair-wise Loss (PPL) function to minimize distances between same-preset-applied photos in latent space. Consequently, the extracted features are robust, enhancing color style transfer for our Deep Preset.

\textbf{Photorealistic Color/Style Transfer (PCT/PST)}. In prior global color transfer works, Reinhard et al. \cite{reinhard2001color} first proposed a low-level computational color transfer method by matching mean and standard deviation. Afterward, Pitie et al. made many efforts in automated color transfer \cite{pitie2005n, pitie2005towards, pitie2007automated} and introduced a transformation based on Monge-Kantorovitch Linear (MKL) \cite{pitie2007linear}. Nowadays, these traditional works are still useful in other fields, such as creating an image harmonization dataset \cite{DoveNet2020}. Adopting the success of deep learning techniques, Gatys et al. \cite{gatys2016image} present an optimization-based method Neural Style Transfer (NST) transferring an artistic style into a photo using convolutional neural networks. Thanks to the surprising performance of NST, the style transfer field gains huge attention from researchers around the world afterward and is growing rapidly. For example, Johnson et al. \cite{johnson2016perceptual} achieve real-time performance in style transfer using a feed-forward deep neural network. Huang et al. \cite{huang2017arbitrary} create a novel way to transform contextual features based on mean and standard deviation (AdaIN); meanwhile, Li et al. \cite{li2017universal} apply whitening and coloring (WCT) for features transform. However, the mentioned methods are designed for artistic stylization rather than photorealistic stylization, which requires high performance in retaining structural details. Therefore, Luan et al. \cite{luan2017deep} propose a regularization for NST to prevent distortion. However, the optimization-based method costs long computational time, and their result is still distorted. Li et al. \cite{li2018closed} propose an enhanced photo stylization PhotoWCT based on WCT with post-processing such as smoothing and filtering techniques. Based on PhotoWCT, Yoo et al. \cite{yoo2019photorealistic} present a progressive strategy transferring style in a single pass and propose Wavelet Corrected Transfer (WCT$^2$) with wavelet pooling/unpooling. Furthermore, they do not need any post-processing; however, their performance still relies on semantic masks. Recently, An et al. \cite{an2019ultrafast} propose asymmetric auto-encoder PhotoNAS without requiring post-processing and guided masks. Their novelty includes two modules, Bottleneck Feature Aggregation (BFA), Instance Normalized Skip Link (INSL), and Network Architecture Search (NAS) for optimizing network architecture under complexity constraint.
However, in blending and retouching photos, the previous methods are overused due to transferring exact colors with degradation rather than learn the color transformation/style representation, which also means "\textit{what beautifies the reference}". Consequently, their results show the content-mismatched colors, hard-copied colors from the reference, and distortion, as shown in Figure~\ref{fig:story}. Meanwhile, the end-users desire to have a homologous color style as a well-retouched reference, especially in sensitive cases such as portraits. In this work, we define that the color style is a preset of low-level image transformation operations converting a photo with natural colors (being closest to what humans actually observe) to retouched ones.
Based on that definition, we present a novel training scheme for color style transfer \textit{with ground-truth} by leveraging various user-generated presets. As a result, having ground-truth helps our model converge in the right direction rather than based on extracted features and transform methods. Furthermore, we propose the Deep Preset to 1) learn well-generalized features representing color style transforming the input (natural) to reference (retouched), 2) estimate the preset applied on the reference, 3) synthesize the well-retouched input. Please check our supplemental document for the visual correlation between photos retouched by the same preset.

\begin{figure*}[htbp]
 \centering
 \includegraphics[width=0.8\textwidth]{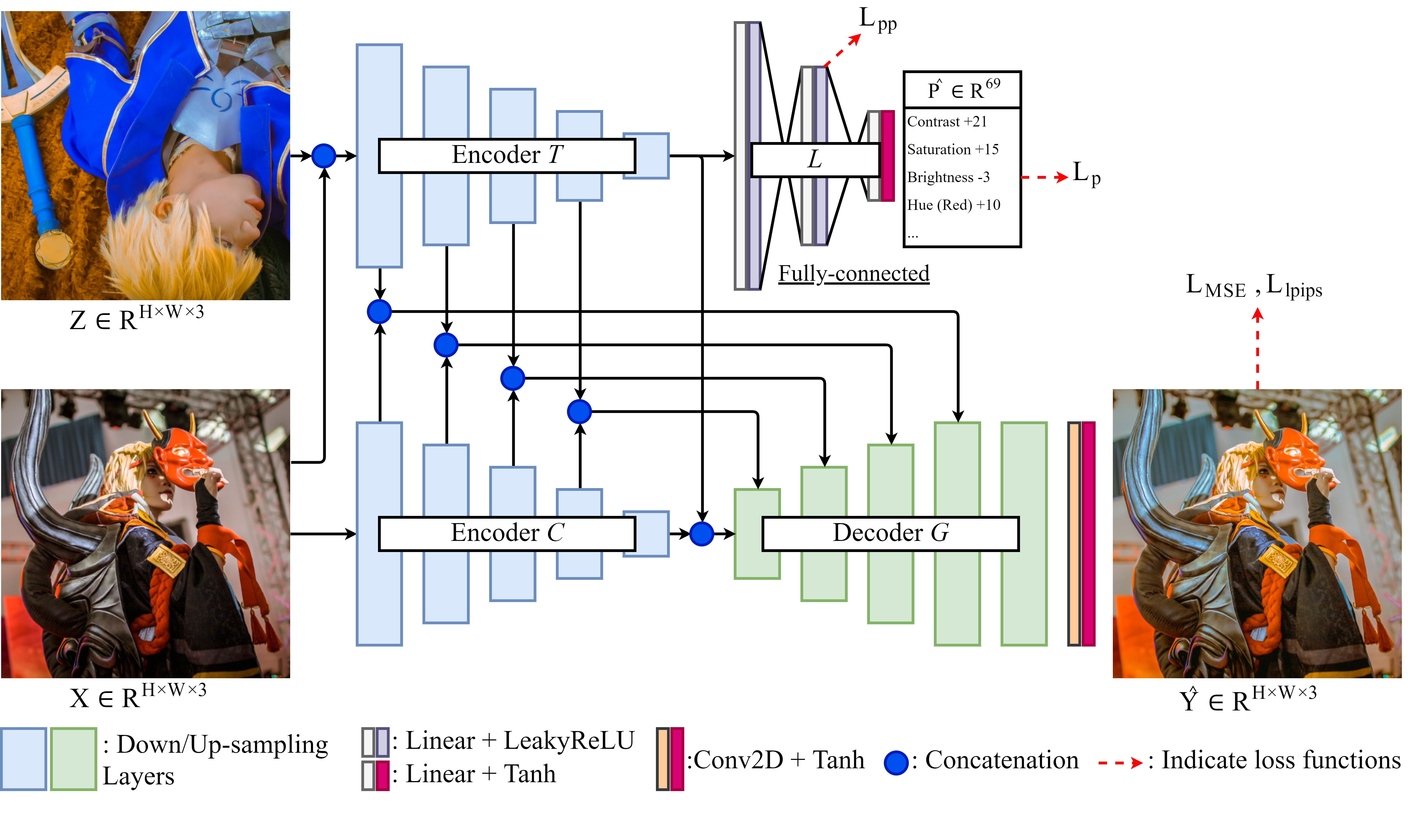}
 \caption{Deep Preset transferring the color style of \textit{Reference} $Z$ to \textit{Input} $X$.}
 \label{fig:dp}
\end{figure*}

Our contributions are as follows:
\begin{itemize}
 \item We present a supervised approach for color style transfer by exploiting various user-generated presets in blending and retouching photos. Furthermore, as a conducted user study, two different images applied by a preset are recognizable.
 \item We propose a specific deep neural network, named Deep Preset, to transfer the color style of a reference to a photo and predict low-level image transformation behind. As a result, our work outperforms previous works in transferring color style qualitatively and quantitatively.
 \item Our Positive Pair-wise Loss (PPL), optimizing the distances between latent features of the photos applied by the same preset, shows the capability to stabilize color transformation and enhance stylized output.
  \item Our work can automatically beautify a photo by selecting its suitable reference among well-retouched photos based on the perceptual measurement \cite{zhang2018unreasonable}.
\end{itemize}

\section{Deep Preset}
\subsection{Overview}
Blending and retouching photos help everyone to enhance the preciousness of their life's moments captured in photographs. Color style is the way of telling expression. However, it is not easy for them to create a plausible color style for their context. They thus search for a well-retouched photo having a similar context for a reference (a). Even a suitable sample is found, it is difficult for the end-users without knowledge in photography to retouch their photos using a powerful photo editing application (b).
In our work, we solve (b) using our proposed Deep Preset, which can synthesize a similar color style from the reference for a specific photo. Additionally, our Deep Preset considers which image transformation operations (preset) have been applied to the reference and learns the features representing the color transformation from natural colors (input) to retouched ones (reference) for different image content.
Regarding the problem (a), we also provide a strategy to find a reference in many well-retouched photos by matching the contextual information \cite{zhang2018unreasonable}. Consequently, the end-users can retouch their photos in one-click. Additionally, we minimize the distance between photos with a similar color transformation in latent space to enhance the generated color style and stabilize preset estimation. Our performance is thus improved.

Our Deep Preset learns the color transformation from a natural photo $X \in \mathbb{R}^{H \times W \times 3}$ to reference $Z \in \mathbb{R}^{H \times W \times 3}$ and generates the stylized photo $\hat{Y} \in \mathbb{R}^{H \times W \times 3}$. Furthermore, our work also predicts the applied preset $P \in \mathbb{R}^{69}$ representing the hyper-parameters of $69$ low-level image transformation operations retouching $Z$, as shown in Figure \ref{fig:dp}. Besides, we extract embeddings $F_{Z}$ and $F_{Z'}$, $\forall F_{Z}, F_{Z'} \in \mathbb{R}^{1024}$ from $Z$ and $Z'$ while predicting the applied preset, where $Z'$ is the random photo retouched by $P$ same as $Z$. Please check our supplemental document for the illustration of how $Z$ and $Z'$ are selected and processed while training.


Our advantages are as follows:
\begin{itemize}
 \item Our models can be converged in the right direction with ground-truth. Meanwhile, previous works are mostly based on feature-based transform techniques.
 \item Learning the low-level image transformation, rather than transferring/mapping exact colors, can reduce the sensitiveness of color style transfer caused by mismatched image content.
  \item Our Positive Pair-wise Loss (PPL) function makes preset estimation stable and enhances generated color style.
\end{itemize}

\subsection{Network Architecture}
We adopt the U-Net \cite{ronneberger2015u}, which has an encoder-decoder architecture, to design the Deep Preset.
Our network includes four main components: Encoder $T$, Encoder $C$, Linear Layers $L$ and Decoder $G$.

First of all, the encoder $T$ leverages the content $X$ and the reference $Z$ to synthesize feature maps representing color transformation. Meanwhile, the encoder $C$ extracts contextual information preparing for blending features between $T$ and $C$. Afterwards, the linear $L$ leverages the final feature map of $T$ to extract the transformation embedding $F_{*}$ and estimate the preset $P$, as follows:

\begin{equation}
 F_{*}, \hat{P} = L(T(X,Z))
 \label{eq_p}
\end{equation}

where $*$ can be $Z$ or $Z'$, $\hat{P}$ is the estimated preset. Finally, the generator $G$ leverages the concatenated features between $T$ and $C$ to synthesize the stylized photo $\hat{Y}$, as:

\begin{equation}
 \hat{Y} = G(T(X,Z) \bullet C(X))
\end{equation}
where $\bullet$ represents concatenations of extracted feature maps between $T$ and $C$ corresponding to feeding order, as shown in Figure \ref{fig:dp}. Please check our supplemental document for the network's technical details.


\subsection{Loss functions}
In this work, we propose a new scheme to train color style transfer with ground-truth; therefore, our loss functions are based on the ground-truth rather than extracted features of content and reference images. Consequently, our models can be converged the right way to be closer to the ground-truth. We apply Mean Square Error (MSE) to directly minimize the distance between our stylized $\hat{Y}$ and the ground-truth $Y$ as:

\begin{equation}
 \mathcal{L}_{MSE} = \frac{1}{N} \sum_{i=1}^{N} {|| Y_i - \hat{Y}_i ||}^2_2
\end{equation}

where N is the batch size. Additionally, we adopt the perceptual loss LPIPS \cite{zhang2018unreasonable} to enhance contextual details as:

\begin{equation}
 \mathcal{L}_{lpips} = LPIPS(\hat{Y},Y)
\end{equation}

Besides, we also predict the preset applying to the $Z$. The estimated preset $\hat{P}$ is observed as:

\begin{equation}
 \mathcal{L}_{p} = \frac{1}{N} \sum_{i=1}^{N} {||P_i - \hat{P}_i||}_{1}
\end{equation}

where $P$ is the number of hyper-parameters representing low-level image transformation methods such as color-shifting. However, predicting an exact preset is difficult due to the variety of possible adjustments. It may be influenced by different image content leading to degrading our stability. We expect that the well-trained model can provide similar features representing a color transformation (preset) for all photos retouched by the same preset. Therefore, in the training stage, we randomly select a photo $Z'$ which is also retouched by $P$, extract the embedding $F_{Z'}$ (as described in the Equation \ref{eq_p}), and finally minimize the error between $F_{Z}$ and $F_{Z'}$, so-called positive pair-wise error, as:

\begin{equation}
 \mathcal{L}_{pp} = \frac{1}{N} \sum_{i=1}^{N} {||{F_{Z'}}_i - {F_{Z}}_i||}_{1}
\end{equation}

Finally, our total loss function is:

\begin{equation}
 \mathcal{L}_{total} = \alpha\mathcal{L}_{MSE} + \beta\mathcal{L}_{lpips} + \gamma\mathcal{L}_{p} + \eta\mathcal{L}_{pp}
\end{equation}

where $\alpha, \beta, \gamma, \eta$ are empirically set as $1, 0.5, 0.01, 1$ respectively. Please check our supplemental document for the illustration of the preset prediction and the positive pair-wise loss function.

\begin{figure*}[t]
 \centering
 \includegraphics[width=0.8\textwidth]{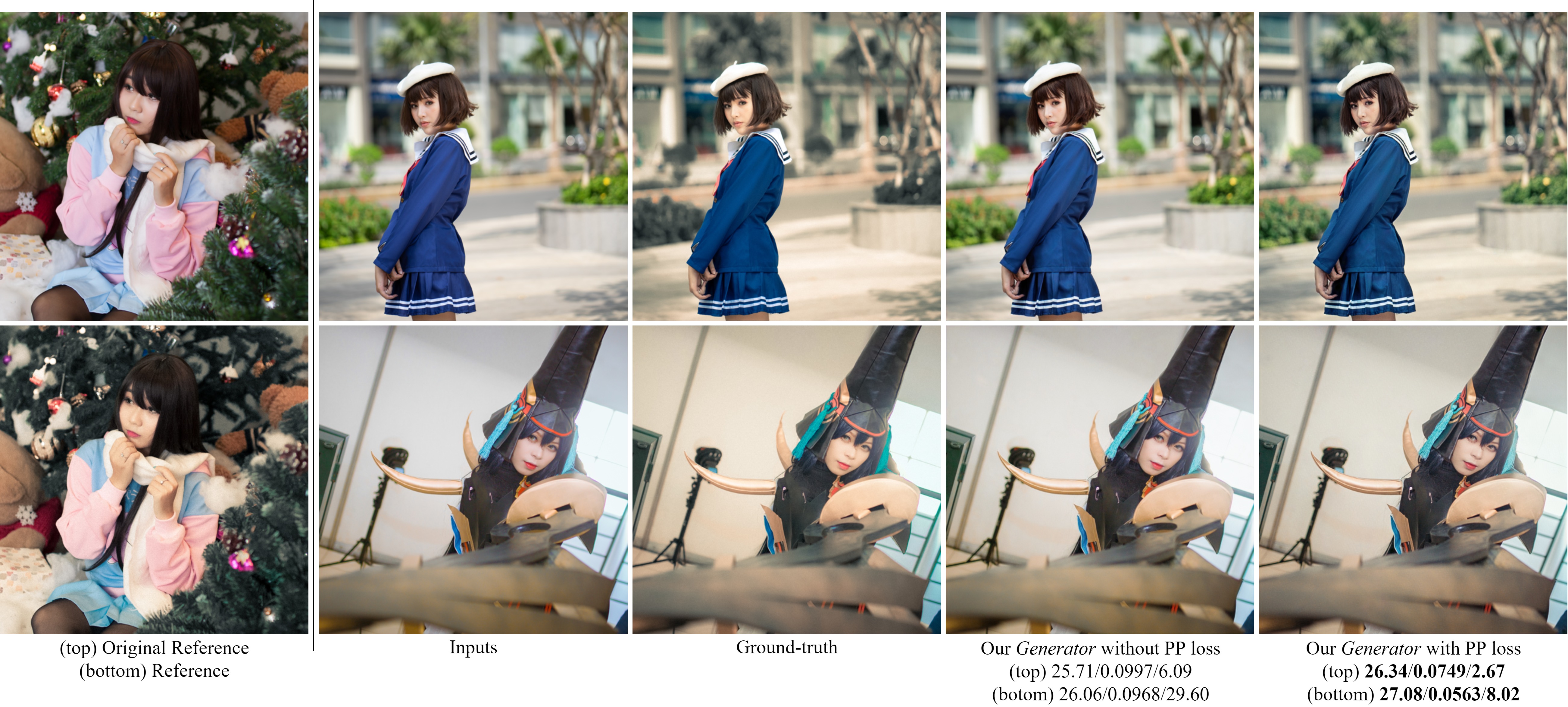}
 \caption{Ablation study on the Positive Pair-wise Loss (PPL) function for our \textit{Generator G}. Training with PP loss achieves the higher performance. Results are measured using PSNR/LPIPS/Chi-squared distances. \textbf{Bold} values mean better.}
 \label{fig:ab}
\end{figure*}

\begin{figure}[t]
 \centering
 \includegraphics[width=0.8\linewidth]{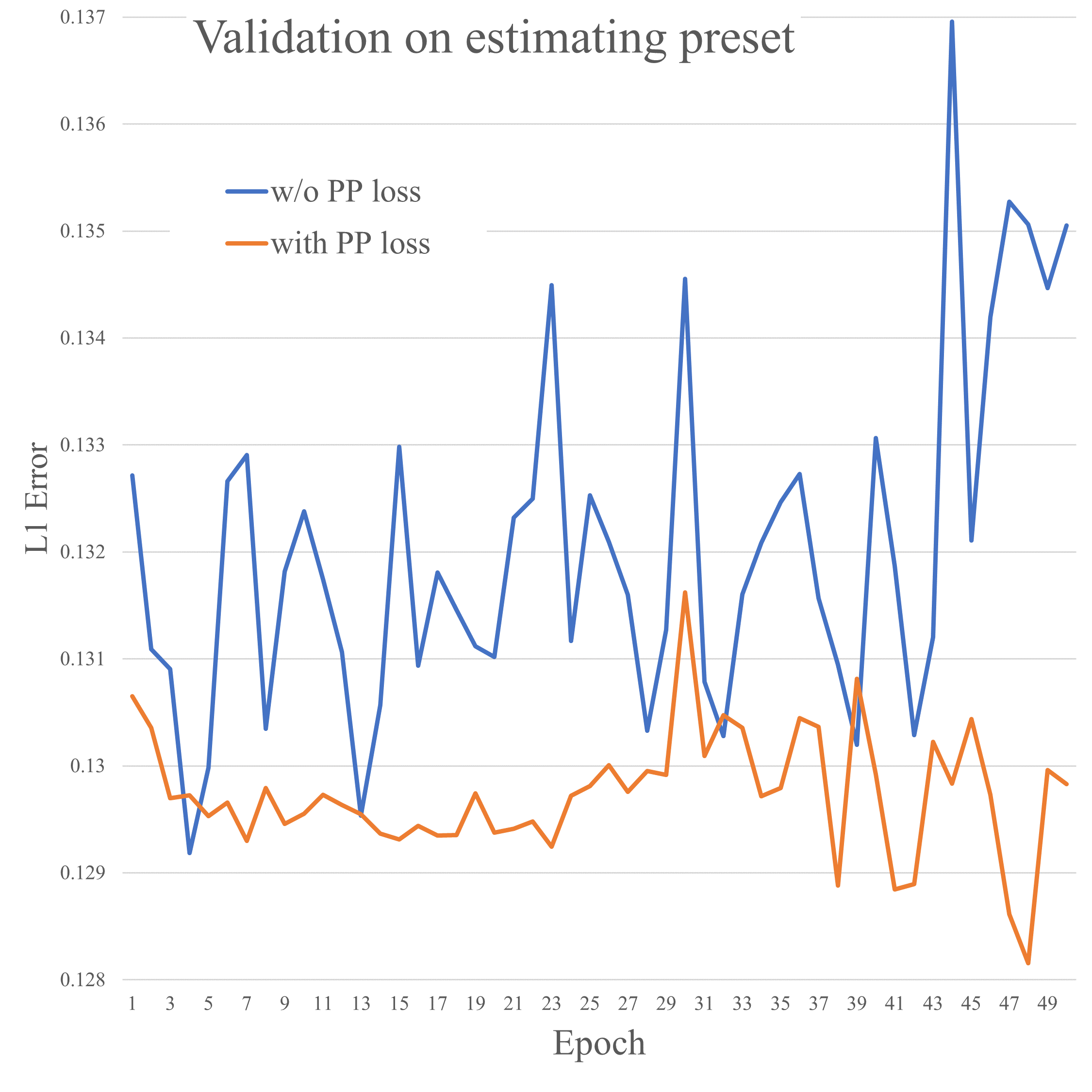}
 \caption{Validation on preset prediction with/without our Positive Pair-wise Loss (PPL) function. Predicted presets with PP loss show more stable.}
 \label{fig:pp}
\end{figure}

\subsection{Data Preparation}
\label{sec:data}
In this section, we describe how we collect and pre-process the data for training and evaluation.

\textbf{Lightroom presets}.
Our data processing is mostly based on Adobe Lightroom \cite{lr}, the most powerful photo editing software recently. We collect 510 user-generated presets, 500 presets for training, and 10 for testing. Additionally, we only select 69 settings (low-level color transformation operations). Each setting has a value representing how large colors are shifted in a specific way. Therefore, a preset with 69 settings is assigned as a 69-dimension vector. All elements are normalized to $[-1, 1]$ based on its min/max values that the end-users can intervene.

\textbf{Training data}.
We script the Lightroom \cite{lr} to generate $601,200$ photos using $1,200$ high-definition photos from Flickr2K \cite{Lim_2017_CVPR_Workshops} and 501 pre-processed presets including base color style. Since our training target is to convert a photo with correct colors (natural) to a retouched version, we only choose the photos having the natural-looking, likely the original colors taken by a camera and corrected by humans. All training photos are scaled to $720$ with the same ratio and compressed by JPEG for efficient storage.

\textbf{Testing data}.
We prepare four subsets for evaluation from DIV2K \cite{Timofte_2018_CVPR_Workshops}, MIT-Adobe FiveK \cite{bychkovsky2011learning}, and Cosplay Portraits (CP) representing the sensitive case in color style transfer. For a fair comparison to the previous works, test data also includes the retouched photos (check our supplemental document for more detail). Regarding the color style, we utilize $10$ user-generated presets to stylize the references. In a particular way, from the DIV2K \cite{Timofte_2018_CVPR_Workshops} validation set, we prepare 1 content image to be stylized by 100 reference images (1x100x10) for the concept of different references stylizing the same image content, and 10 content images and 10 reference images (10x10x10). Taking into account of having various contexts in testing, we randomly select a set of 10x10x10 including all contextual categories such as \textit{indoor}, \textit{outdoor}, \textit{day}, \textit{sun sky}, etc. from MIT-Adobe FiveK \cite{bychkovsky2011learning}. Since our work is color transfer instead of color enhancement; plus, the raw images from \cite{bychkovsky2011learning} are not always taken and processed well as our expectation of naturalness; we thus choose the images retouched by the \textit{Expert C} only. Regarding test images of CP, we choose a set of 10x10x10 including both natural colors originally taken by modern DSLR cameras and retouched ones. In summary, we evaluate all methods on $1000$ samples each set and $4000$ samples in total. All photos are stored in JPEG and resized to $512 \times 512$ on-the-fly using bicubic interpolation while testing.

\begin{figure*}[t]
 \centering
 \includegraphics[width=\textwidth]{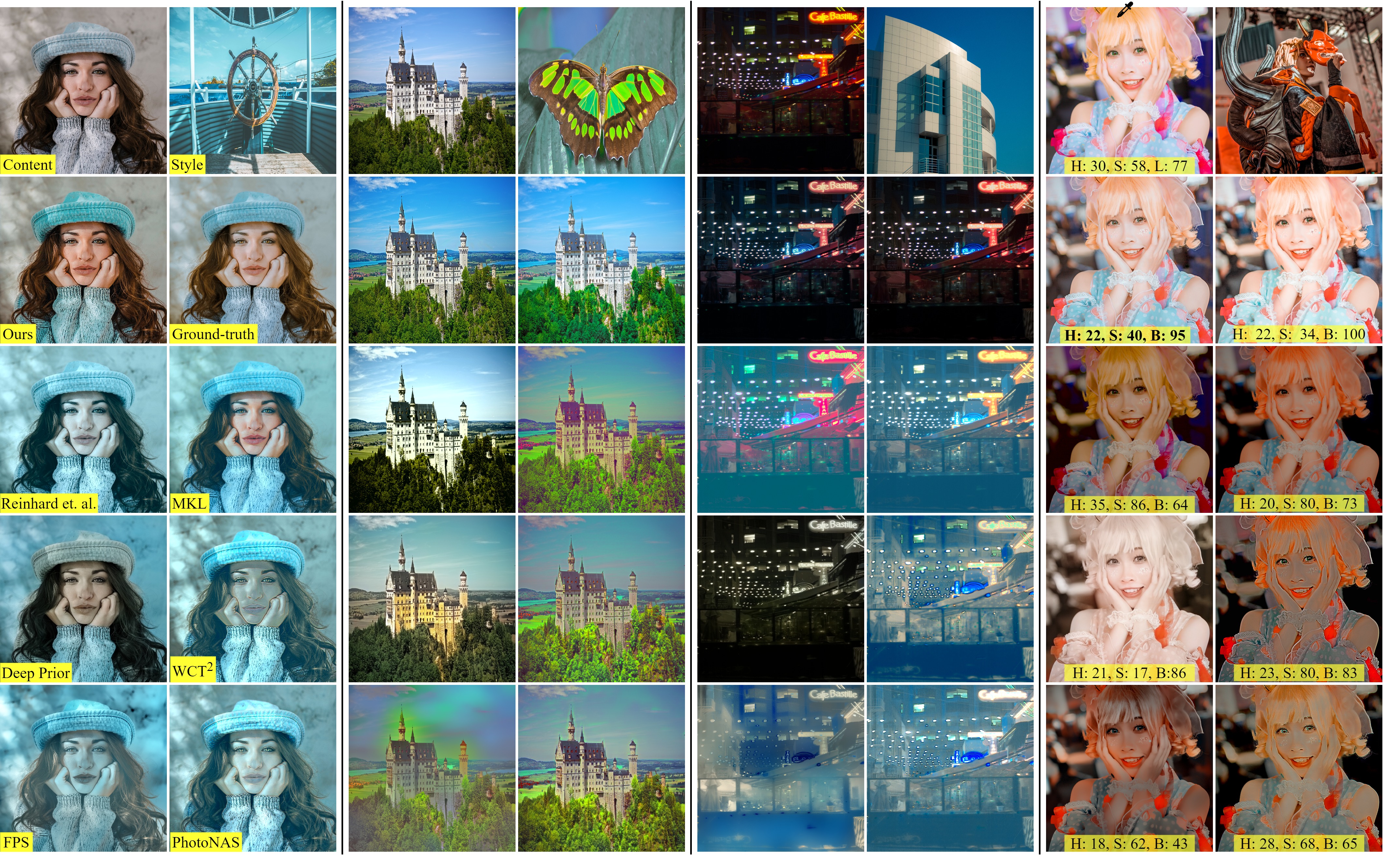}
 \caption{Qualitative comparison between our Deep Preset (\textit{Generator}) and the previous works Reinhard et al. \cite{reinhard2001color}, MKL \cite{pitie2007linear}, Deep Priors \cite{zhang2017real}, FPS \cite{li2018closed}, WCT$^2$ \cite{yoo2019photorealistic}, PhotoNAS \cite{an2019ultrafast}. \textit{H, S, B} denotes Hue, Saturation, and Brightness, respectively. \textit{Left-to-right}: DIV2K \cite{Timofte_2018_CVPR_Workshops} 1x100x10, DIV2K \cite{Timofte_2018_CVPR_Workshops} 10x10x10, Adobe-MIT FiveK \cite{bychkovsky2011learning} 10x10x10, and Cosplay Portraits 10x10x10.}
 \label{fig:comparison}
\end{figure*}

\begin{table*}[t]
\caption{Quantitative comparison between our models with/without Positive-Pairwise Loss (PPL), the previous works Deep Priors \cite{zhang2017real}, Fast Photo Style Transfer (FPS) \cite{li2018closed}, WCT$^2$ \cite{yoo2019photorealistic}, and PhotoNAS \cite{an2019ultrafast}. \textbf{Bold} values indicate the best performance. $\uparrow$: higher is better, $\downarrow$: lower is better. We choose our \textit{Generator*} for other comparisons throughout the paper.}
\label{tab:comparison}
\resizebox{\textwidth}{!}{%
\begin{tabular}{|l|l|c|c|c|c|c|c|c|c|c|c|c|c|l|l|l|l|}
\hline
\multicolumn{2}{|c|}{\multirow{2}{*}{Method}} & \multicolumn{4}{c|}{DIV2K (1x100x10 and 10x10x10)} & \multicolumn{4}{c|}{MIT-Adobe FiveK (10x10x10)} & \multicolumn{4}{c|}{Cosplay Portraits (10x10x10)} & \multicolumn{4}{c|}{Average} \\ \cline{3-18} 
\multicolumn{2}{|c|}{} & H-Corr $\uparrow$ & H-CHI $\downarrow$ & PSNR $\uparrow$ & LPIPS $\downarrow$ & H-Corr $\uparrow$ & H-CHI $\downarrow$ & PSNR $\uparrow$ & LPIPS $\downarrow$ & H-Corr $\uparrow$ & H-CHI $\downarrow$ & PSNR $\uparrow$ & LPIPS $\downarrow$ & \multicolumn{1}{c|}{H-Corr $\uparrow$} & \multicolumn{1}{c|}{H-CHI $\downarrow$} & \multicolumn{1}{c|}{PSNR $\uparrow$} & \multicolumn{1}{c|}{LPIPS $\downarrow$} \\ \hline \hline
\multicolumn{2}{|l|}{Reinhard et. al. \cite{reinhard2001color}} & 0.3069 & 915.63 & 15.77 & 0.2620 & 0.2198 & 969.67 & 13.62 & 0.3222 & 0.2341 & 1104.06 & 14.03 & 0.2764 & 0.2536 & 996.45 & 14.47 & 0.2869 \\ \hline
\multicolumn{2}{|l|}{MKL \cite{pitie2007linear}} & 0.3390 & 662.45 & 16.20 & 0.2607 & 0.1785 & 819.37 & 13.59 & 0.3151 & 0.3037 & 566.81 & 14.41 & 0.2545 & 0.2737 & 682.88 & 14.73 & 0.2768 \\ \hline
\multicolumn{2}{|l|}{Deep Priors \cite{zhang2017real}} & 0.5420 & 749.50 & 20.53 & 0.2033 & 0.4049 & 785.87 & 16.84 & 0.2947 & 0.4735 & 656.08 & 17.68 & 0.2896 & 0.4735 & 730.49 & 18.35 & 0.2625 \\ \hline
\multicolumn{2}{|l|}{FPS} & 0.3856 & 1232.97 & 14.71 & 0.3025 & 0.1800 & 1843.86 & 12.05 & 0.3902 & 0.3363 & 1629.30 & 12.93 & 0.3105 & 0.3006 & 1568.71 & 13.23 & 0.3344 \\ \hline
\multicolumn{2}{|l|}{$\text{WCT}^2$ \cite{yoo2019photorealistic}} & 0.3917 & 1269.91 & 16.40 & 0.2726 & 0.2043 & 5916.76 & 13.18 & 0.3633 & 0.3201 & 2950.11 & 13.98 & 0.2775 & 0.3054 & 3378.92 & 14.52 & 0.3045 \\ \hline
\multicolumn{2}{|l|}{PhotoNAS \cite{an2019ultrafast}} & 0.4129 & 824.74 & 17.06 & 0.2559 & 0.1898 & 7924.73 & 13.51 & 0.3849 & 0.3119 & 3853.21 & 14.29 & 0.2975 & 0.3049 & 4200.89 & 14.95 & 0.3128 \\ \hline
\multicolumn{1}{|c|}{\multirow{2}{*}{Ours w/o PPL}} & Preset Prediction & 0.6416 & 509.71 & 22.62 & 0.1139 & 0.6094 & 673.36 & 21.07 & 0.1179 & 0.6569 & 389.60 & 21.88 & 0.1042 & 0.6360 & 524.23 & 21.86 & 0.1120 \\ \cline{2-18} 
\multicolumn{1}{|c|}{} & Generator & 0.6933 & 558.56 & 22.87 & 0.1027 & \textbf{0.6415} & 454.99 & \textbf{21.86} & 0.1105 & 0.6713 & 309.17 & 21.78 & 0.0992 & 0.6687 & 440.91 & 22.17 & 0.1041 \\ \hline
\multicolumn{1}{|c|}{\multirow{2}{*}{Ours w PPL}} & Preset Prediction & 0.6194 & \textbf{299.34} & 22.03 & 0.1275 & 0.6276 & \textbf{319.79} & 20.98 & 0.1160 & 0.6222 & \textbf{258.67} & 21.44 & 0.1157 & 0.6231 & \textbf{292.60} & 21.48 & 0.1197 \\ \cline{2-18} 
\multicolumn{1}{|c|}{} & Generator* & \textbf{0.7006} & 552.45 & \textbf{23.12} & \textbf{0.0980} & 0.6313 & 968.51 & 21.71 & \textbf{0.1093} & \textbf{0.6927} & 325.89 & \textbf{22.15} & \textbf{0.0960} & \textbf{0.6749} & 615.62 & \textbf{22.33} & \textbf{0.1011} \\ \hline
\end{tabular}%
}
\end{table*}


\section{Experimental results}

\subsection{On our Positive Pair-wise Loss (PPL) function}
The encoder $T$ of Deep Preset learns color transformation representation with an auxiliary regression as the preset prediction. However, it is difficult to estimate an accurate preset leading to the instability of the extracted features representing a specific transformation with different image contents.
Therefore, we consolidate the color transformation by optimizing distances between photos having the same color style in latent space. The extracted features are thus robust for transformation making preset prediction stable while training. To prove it, we train two models with/without PPL function in the same condition and compare them qualitatively and quantitatively using histogram correlation (H-Corr), histogram Chi-squared (H-CHI), Peak Signal-to-Noise Ratio (PSNR), and the perceptual metric LPIPS \cite{zhang2018unreasonable}. Regarding evaluating the preset prediction, instead of calculating an error between predicted presets $\hat{P}$ and the actual presets $P$, we apply the $\hat{P}$ back to the content images to achieve the stylized photos so that our preset prediction can be further compared to the previous works.
As a result, in predicting presets, the model trained without PPL (non-PPL model) outperforms the model trained with PPL (PPL model) with higher H-Corr as \textbf{0.6360}, higher PSNR as \textbf{21.86} dB, lower LPIPS as \textbf{0.1120}. However, regarding directly generating the output by the Generator \textit{G}, the PP model quantitatively outperforms the non-PPL model with higher H-Corr as \textbf{0.6749}, higher PSNR as \textbf{22.33} dB, lower LPIPS as \textbf{0.1011}, as shown in Table \ref{tab:comparison}. Additionally, the PPL model stabilizes the preset prediction under different image contents for a specific color style, as shown in Figure \ref{fig:pp}. Qualitatively, the PPL model gives the closer color style as \textit{yellow} tone to the reference with higher PSNR, lower LPIPS showing a better quality; furthermore, it has smaller histogram-related distances proving the PP model's outperformance in color transformation, as shown in Figure \ref{fig:ab}. Please check our supplemental document for a qualitative comparison of both models' preset prediction's stability and further discussion on the trade-offs between preset prediction and PPL.

\subsection{Comparison to recent works}
In this section, we compare our work to Reinhard et. al. \cite{reinhard2001color}, Monge-Kantorovitch Linear (MKL) \cite{pitie2007linear}, Fast Photo Style Transfer \cite{li2018closed}, WCT$^2$ \cite{yoo2019photorealistic}, PhotoNAS \cite{an2019ultrafast} presenting creative techniques in photorealistic color/style transfer. Their works show the capability of transferring textures and colors of reference into another photo, even changing the day to night, summer to winter. However, they are overused in blending and retouching photos. This work defines that the color style is made by the low-level image transformation (e.g., color-shifting) converting a photo with natural colors to its retouched version. Adopting that idea, our proposed Deep Preset learns the defined color style representation and transforms the base colors with ground-truth instead of transferring exact colors from the reference. Hence, our work does not degrade the image quality but can beautify the input images based on a reference. To prove our proficiency, we compare our work to the mentioned works in quantitative and qualitative ways. Besides, the works in colorization with reference-guided can be treated as color style transfer. Therefore, we also compare this work to the interactive colorization of Zhang et al. \cite{zhang2017real}. Since our scheme provides the ground-truth, we thus utilize the previously mentioned similarity metrics such as H-Corr, H-CHI, PSNR, and LPIPS \cite{zhang2018unreasonable} for quantitative comparison on the four subsets described in Section \ref{sec:data}.  Furthermore, we qualitatively show a sample from each subset in various contexts to convince our quantitative result. Considering this work under the aspects of production, we conduct a user study based on two-alternative forced-choice (2AFC) measuring human perceptual similarity and user preferences.


\textbf{Quantitative comparison}.
We compare our work to previous works on four subsets consisting of DIV2K \cite{Timofte_2018_CVPR_Workshops} 1x100x10, DIV2K \cite{Timofte_2018_CVPR_Workshops} 10x10x10, MIT-Adobe FiveK \cite{bychkovsky2011learning} 10x10x10, and Cosplay Portraits 10x10x10, as described in Section \ref{sec:data}. As a result, the reference-guided colorization Deep Priors \cite{zhang2017real} outperforms other previous works on average. Their colorization removes color channels, then colorize the black-and-white photo based on a given reference. The generated color thus has a high correlation with the ground-truth. However, they still suffer from color overflowed and mismatched. Meanwhile, this work quantitatively outperforms the previous works in generating a similar color style with a higher H-Corr as \textbf{0.6749} and H-CHI \textbf{615.62}. Furthermore, our results also achieve the highest PSNR as \textbf{22.33} dB, lowest LPIPS as \textbf{0.1011} on average, as shown in Table \ref{tab:comparison}.

\textbf{Qualitative comparison}.
To support our quantitative comparison, we show four samples from four test subsets described in Section \ref{sec:data}. As a qualitative result, the proposed Deep Preset can beautify the content image using a well-retouched reference without reducing image quality; meanwhile, the previous works try to transfer exact colors leading to abnormal colors causing unnaturalness. Particularly, the previous works transfer the \textit{globally bluish tone} of reference to the whole content image losing the color of \textit{the girl's hair} in the first two columns. In contrast, our result obtains plausible color for \textit{the skin} with \textit{blonde hair}, which are closest to the ground-truth. Being similar to the following samples, our Deep Preset provides a plausible color in harmony regardless of image content; furthermore, our results have the most similar color style to the ground-truth. For example, the \textit{sky} turns \textit{bluer} and the \textit{trees} turn \textit{brightly greener} in the second sample, the saturation of the text "\textit{Cafe Bastille}" in the third sample is reduced being similar to the ground-truth, even the context between the content image and reference is mismatched. Meanwhile, the previous works, one way or another, distort the input images by transferring the exact colors of references. In the last sample, we directly check Hue-Saturation-Brightness (HSB) information where the color picker is located. As a result, our work provides the closest values (H,S,B) to the ground-truth as ($\Delta0$, $\Delta-6$, $\Delta5$) compared to \cite{reinhard2001color, pitie2007linear, zhang2017real, li2018closed, yoo2019photorealistic, an2019ultrafast}, as shown in Figure \ref{fig:comparison}. We conclude that our Deep Preset synthesizes the best visual quality, which is essential for production, with the closest color style to the ground-truth compared to the previous works. Please check our supplementary materials for more qualitative comparison.

\begin{table}[t]
\caption{Three user study scenarios based on two-alternative forced-choice (2AFC) selecting A or B anchored by ground-truth, reference (style), and user preferences. Each method is paired with others twice and evaluated by the probability of being chosen (higher is better). \textbf{Bold} values reveal our outperformance compared to the previous works.}
\label{tab:us}
\resizebox{\linewidth}{!}{%
\begin{tabular}{|l|c|c|c|}
\hline
Method & \multicolumn{1}{l|}{\begin{tabular}[c]{@{}l@{}}{[}Triplet{]} Anchored\\ by Grouth-truth\end{tabular}} & \multicolumn{1}{l|}{\begin{tabular}[c]{@{}l@{}}{[}Triplet{]} Anchored\\ by Reference\end{tabular}} & \multicolumn{1}{l|}{\begin{tabular}[c]{@{}l@{}}{[}Pair{]} Anchored\\ by User Preferences\end{tabular}} \\ \hline \hline
Reinhard's \cite{reinhard2001color} & 0.43 & 0.20 & 0.47 \\ \hline
MKL \cite{pitie2007linear} & 0.50 & 0.48 & 0.35 \\ \hline
Deep Priors \cite{zhang2017real} & 0.38 & 0.52 & 0.52 \\ \hline
FPS \cite{li2018closed} & 0.34 & 0.53 & 0.27 \\ \hline
$\text{WCT}^2$ \cite{yoo2019photorealistic} & 0.32 & 0.54 & 0.31 \\ \hline
PhotoNAS \cite{an2019ultrafast} & 0.59 & 0.54 & 0.38 \\ \hline
Ours & \textbf{0.95} & \textbf{0.57} & \textbf{0.85} \\ \hline
Ground-truth & $\infty$ & 0.61 & 0.85 \\ \hline

\end{tabular}%
}
\end{table}

\subsection{User Study}
Our user study includes three scenarios based on the two-alternative forced-choice (2AFC) scheme selecting A or B anchored by (i) ground-truth having the same image content as A and B, (ii) reference having the different image content from A and B, and (iii) user preferences. In the mentioned scenarios, (i) represents the human perceptual similarity metric, which has the same meaning as the histogram distances, PSNR, LPIPS we used in Table \ref{tab:comparison}, (ii) measures whether the human can recognize the similar color style when the content between the input image and reference is different. Meanwhile, (iii) measures how the compared methods are suitable for production. We first guide the users on how the photos retouched by a preset look like, then show examples of the problems in style transfer such as distortion, color overflowed, and more. Afterward, we let the user select picture A or B while asking the question "\textit{Which photo is closest to the anchor?}". Regarding judging based on user preferences, the volunteers are asked, "\textit{Which photo do you like most?}", "\textit{Which photo are you willing to share on your social media?}", and related questions. The samples consisting of triplets for (i) and (ii), pairs for (iii) shown in our experiment are collected randomly from DIV2K \cite{Timofte_2018_CVPR_Workshops} 10x10x10 and Cosplay Portraits 10x10x10 so that each method is paired twice with the other methods; therefore, they are fairly compared. Regarding the users, we invite $23$ volunteers to run the test; in that, \textbf{74\%} of them have experience in photo adjustment. Eventually, they return $3150$ votes for three scenarios in total. As a result in Table \ref{tab:us}, our work has the highest probability of being chosen whenever it is compared to the previous works such as Reinhard's work \cite{reinhard2001color}, MKL \cite{pitie2007linear}, Deep Priors \cite{zhang2017real}, FPS \cite{li2018closed}, WCT$^2$ \cite{yoo2019photorealistic}, PhotoNAS \cite{an2019ultrafast}. Please check our supplemental document for the illustration of our user study.

"\textit{Can human judge color style produced by preset?}" Yes, the human can, as the conducted experiment 2AFC anchored by reference. This scenario will exploit the human perceptual measurement on the defined color style when the image content is not identical. As a result, the ground-truth retouched by the same preset with the reference has the highest number of chosen times as \textbf{61\%}. The result also reveals that a preset can represent a color style. Besides, our Deep Preset provides the most similar color style to the reference as \textbf{57\%} of being chosen compared to the previous works \cite{reinhard2001color, pitie2007linear, zhang2017real, li2018closed, yoo2019photorealistic, an2019ultrafast}, as shown in Table \ref{tab:us}. To have the color style's overall observation, please check our supplemental document for the visual comparison between the photos applied by the same/different preset.

"\textit{Does our Deep Preset degrade the image quality?}" No, it does not. Considering the production aspect, we conduct the 2AFC test anchored by user preferences as our scenario (iii) shown in Table \ref{tab:us}. As a result, our stylized photos achieve the highest probability of being chosen compared to the previous works \cite{reinhard2001color, pitie2007linear, zhang2017real, li2018closed, yoo2019photorealistic, an2019ultrafast}, the same number as the ground-truth. Qualitatively, our work usually provides a satisfying look as content images and ground-truth, as shown in Figure \ref{fig:comparison} and additional results in our supplemental document.

\section{Conclusion}
We define a novel color style based on low-level image transformations from natural to a retouched domain. Adopting that definition, we present a supervised approach for color style transfer, and propose the Deep Preset designed to not only efficiently transfer the color style but also predict the applied preset behind a well-retouched reference. Besides, we present a Positive Pair-wise Loss (PPL) optimizing distances between the photos applied by the same preset in latent space. An ablation study shows that our PPL can stabilize the preset prediction and enhance stylizing. As a result, the proposed Deep Preset outperforms previous works quantitatively and qualitatively. Furthermore, the conducted user study shows that our results achieve the closest color style to the ground-truth, the closest color style to the reference, and the most satisfaction voted by humans.

{\small
\bibliographystyle{ieee_fullname}
\bibliography{citation}
}

\end{document}